\pdfoutput=1

\documentclass[11pt]{article}

\usepackage[preprint]{Style/acl}

\usepackage{times}
\usepackage{latexsym}

\usepackage[T1]{fontenc}

\usepackage[utf8]{inputenc}

\usepackage{microtype}

\usepackage{inconsolata}

\usepackage{microtype}
\usepackage{booktabs}
\usepackage{graphicx}
\usepackage{multirow}
\usepackage{amsmath}
\usepackage{soul}
\usepackage{multicol}
\usepackage{subcaption}
\usepackage{xcolor}

\usepackage[nameinlink,capitalize]{cleveref}

\usepackage{acro}
\newcommand{\na}[3]{\DeclareAcronym{#1}{short=#2, long=#3}}
\na{sr}{SR}{Speech Recognition}
\na{ol}{OL}{Online Learning}
\na{mt}{MT}{Machine Translation}
\na{smt}{SMT}{Statistical Machine Translation}
\na{nmt}{NMT}{Neural Machine Translation}
\na{rnn}{RNN}{Recurrent Neural Network}
\na{ipr}{IPR}{Interactive Pattern Recognition}
\na{mle}{MLE}{Maximum Likelihood Estimation}
\na{bpe}{BPE}{Byte Pair Encoding}
\na{llm}{LLM}{Large Language Model}
\na{sota}{SoTA}{state-of-the-art}
\na{pe}{PE}{Post-Edition}
\na{cat}{CAT}{Computer-Assisted Translation}
\na{imt}{IMT}{Interactive Machine Translation}
\na{ipmt}{IPMT}{Interactive-Predictive Machine Translation}
\na{ipnmt}{IPNMT}{Interactive-Predictive Neural Machine Translation}
\na{ipsmt}{IPSMT}{Interactive-Predictive Statistical Machine Translation}
\na{rl}{RL}{Reinforcement Learning}
 \na{cm}{CM}{Confidence Measures}
 \na{cm2}{CMs}{Confidence Measures}
\na{ma}{MA}{Mouse Action}
\na{ma2}{MAs}{Mouse Actions}
\na{umar}{uMAR}{useful MAR}
\na{cmar}{cMAR}{character MAR}
\na{mar}{MAR}{Mouse Action Ratio}
\na{mac}{MAC}{Mouse Action Count}
\na{csr}{CSR}{Character Stroke Ratio}
\na{csc}{CSC}{Character Stroke Count}
\na{wsr}{WSR}{Word Stroke Ratio}
\na{wsc}{WSC}{Word Stroke Count}
\na{hter}{HTER}{Human-targeted Translation Error Rate}
\na{ter}{TER}{Translation Error Rate}
\na{bleu}{Bleu}{BiLingual Evaluation Understudy}
\na{meteor}{Meteor}{Metric for Evaluation of Translation with Explicit ORdering}
\na{chrf}{Chr-F}{Chracter n-gram F-score}
\na{cer}{CER}{Classification Error Rate}
\na{ksr}{KSR}{Key Stroke Ratio}
\na{mbart}{mBART}{multilingual Bidirectional and Auto-Regressive Transformer}
\na{mt5}{mT5}{multilingual Text-to-Text Transfer Transformer}
\na{xlm}{XLM}{cross-lingual Language Model}

\definecolor{seagreen}{HTML}{2e8b57}
\definecolor{navy}{HTML}{000080}
\newcommand{\val}[1]{\textcolor{seagreen}{\emph{#1}}}
\newcommand{\err}[1]{\textcolor{navy}{\textbf{#1}}}
\newcommand{\argmax}[1]{\underset{\substack{#1}}{\operatorname{arg}\,\operatorname{max}}\;}

\begin{document}

\title{\bf Segment-Based Interactive Machine Translation for Pre-trained Models}

\author{   \bf {\'A}ngel Navarro$^{1}$ \\
                 \texttt{annamar8@prhlt.upv.es} \\
                 \And		  
			\bf Francisco Casacuberta $^{1,2}$ \\
			\texttt{fcn@prhlt.upv.es} \\
			\AND
			\\
			$^{1}$PRHLT, Universitat Politècnica de València, Spain \\
			$^{2}$ValgrAI - Valencian Graduate School and Research Network for Artificial Intelligence, \\
                 Camí de Vera s/n, 46022 Valencia, Spain
}

\maketitle
\pagestyle{empty}

\begin{abstract}
Pre-trained large language models (LLM) are starting to be widely used in many applications. In this work, we explore the use of these models in interactive machine translation (IMT) environments. In particular, we have chosen mBART (multilingual Bidirectional and Auto-Regressive Transformer) and mT5 (multilingual Text-to-Text Transfer Transformer) as the LLMs to perform our experiments. The system generates perfect translations interactively using the feedback provided by the user at each iteration. The Neural Machine Translation (NMT) model generates a preliminary hypothesis with the feedback, and the user validates new correct segments and performs a word correction--repeating the process until the sentence is correctly translated. We compared the performance of mBART, mT5, and a state-of-the-art (SoTA) machine translation model on a benchmark dataset regarding user effort, Word Stroke Ratio (WSR), Key Stroke Ratio (KSR), and Mouse Action Ratio (MAR). The experimental results indicate that mBART performed comparably with SoTA models, suggesting that it is a viable option for this field of IMT. The implications of this finding extend to the development of new machine translation models for interactive environments, as it indicates that some novel pre-trained models exhibit SoTA performance in this domain, highlighting the potential benefits of adapting these models to specific needs.
\end{abstract}

\section{Introduction}
\Ac{mt} has become an indispensable component of contemporary communication, facilitating the interaction among individuals from diverse linguistic backgrounds across borders. Although the quality of the translations generated by the \ac{mt} systems has improved highly in recent years thanks to the apparition of \ac{nmt}, the systems are not perfect yet, and there are a large set of variables that can alter the quality of the translations generated, as the data used, or the training method \citep{Xu24,Choshen19}. Notably, pre-trained multilingual \ac{mt} models, such as \ac{mbart}  \citep{Liu20}, \ac{mt5} \citep{Xue20}, and \ac{xlm} \citep{Lample19}, have made a significant advancement in the \ac{mt} field, as these models demonstrate state-of-the-art performance across various benchmark datasets. Now, we can obtain high-quality translations for a specific task or domain by fine-tuning them with a non-extensive training data set.

Although we can obtain high-quality translations by fine-tuning these pre-trained models, there are still challenges in achieving accurate and fluent translations for all languages and domains \citep{Toral20}. Moreover, there are domains such as medical or judicial where a translation error could have a significant consequence, so it is mandatory to have a human translator to check or post-edit these translations. To reduce the effort performed during the post-edition process and to ensure perfect translations, the \ac{imt} field appears \citep{Peris17,Alabau13b,Koehn09}. \ac{imt} systems combine interactively an \ac{mt} model with a human translator. The system allows the user to interactively participate in the translation process by providing feedback to the machine at each iteration, which generates a new translation that improves the previous one. This process is repeated until the system generates a perfect translation and the user validates it.

This paper explores the performance of different pre-trained multilingual \ac{mt} models in an \ac{imt} environment. We aim to compare the effectiveness of these models that can assess state-of-the-art results in a large set of translation tasks in the field of IMT. Although both tasks are very similar, \ac{imt} needs the \ac{mt} model used for generating the translations to adapt to the feedback the user is sending; it needs a good generalization capability to think of a new good translation that fulfills the user feedback. \citet{Navarro23a} already compared \ac{mbart} with a system trained from scratch using the prefix-based protocol \citep{Foster97}. In this paper, we use the segment-based protocol \citep{Domingo17,Peris17}, which is more similar to post-edition, and also compare the results with a new pre-trained multilingual \ac{mt} model. To achieve this, we designed and implemented an \ac{imt} system with a segment-based protocol that can integrate different models from the Hugging Face library \citep{Wolf20}. In this protocol, starting with the system generating a hypothesis, the user must validate all the correct word segments and then perform a word correction. This process repeats until the user validates the translation generated. To compare the models, we use two subsets of metrics: a first one to know the quality of the first hypothesis generated and a second set of metrics to measure the human effort performed during the translation session. Our results showed that although both models have good quality results for the first hypothesis, their generalization capability is not as good as state-of-the-art \ac{mt} models. mBART has obtained a higher Bleu score, but as the model struggles to generalize the content between the validated segments, the number of iterations needed to finish the translation process is higher than the state-of-the-art. Based on our findings, it can be concluded that while pre-trained models generally achieve better translation quality than our models trained from scratch, they cannot generalize as well when used in an \ac{imt} environment at the segment level.

Our work presents several contributions to the field of \ac{imt}. Specifically, our contributions are as follows:
\begin{itemize}
    \item \textbf{Creation of an \ac{imt} system at the segment level}: We have implemented an \ac{imt} system with a segment-based protocol that can use different \ac{mt} models. Let the system generate the first hypothesis and then send simulated user feedback to the model to obtain better translations. This process is repeated until the system translates correctly the source sentence.
    \item \textbf{Compare IMT results beetween pre-trained model}: We analyze and compare the results obtained with the pre-trained models. Our study evaluates the quality of the translations and the level of user effort required during translation sessions.
    \item \textbf{Compare IMT results with traditional techniques}: We analyze and compare the results obtained with the pre-trained models with one trained from scratch using the OpenNMT-py toolkit. Our study evaluates the quality of the translations and the level of user effort required during translation sessions.
\end{itemize}

The rest of the paper is organized as follows. In Section 2, we provide a brief overview of related work in pre-trained language models and \ac{imt}. Section 3 describes our proposed approach in detail, including the architecture of the \ac{imt} system and the feedback mechanism. In Section 4, we present the experimental framework, and in Section 5, we discuss the results obtained. Finally, we conclude the paper in Section 6 and discuss potential future directions for this research.

\section{Related Work}
In this paper, our central focus is to examine the effectiveness of pre-trained \ac{mt} models in our specific task, segment-based \ac{imt}. Training an \ac{mt} model from scratch for a specific translation task or domain can be time-consuming and expensive, so it will be very beneficial if we know in advance if pre-trained models can also obtain good results in this scenario. These models have successfully proven that they can obtain state-of-the-art results in various tasks, such as translation, but although \ac{mt} and \ac{imt} tasks are similar, the goal of the first is to obtain the most accurate translation, while the second tries to obtain a perfect translation with minimal user interaction. For the second case, we need the model to adapt to the user feedback, generating a reasonable hypothesis each time that fulfills the validated segments.

This paper compares the pre-trained multilingual models mBART and mT5 to each other and compares them with the performance of a state-of-the-art \ac{nmt} model trained from scratch for the specific task. There are other pre-trained multilingual models, such as XLM, \citep{Lample19} DeltaLM \citep{Ma21}, or XGLM \citep{Lin21}, that we could have used for our purpose. As we have used, other people are trying to use these models for new tasks that were not initially thought. \cite{Shen21} used to resolve math word problems, \cite{Farahani21} to summarize Persian texts, \cite{Chakrabarty21} generated poetry with them, and \cite{Li20} implemented it in the speech translation field.

Beyond pre-trained multilingual models, the emergence of \ac{llm} has spurred investigations into their application for specialized tasks, such as translation. These models have undergone rigorous training on extensive multilingual datasets, equipping them with the capability to capture intricate linguistic patterns and translations across a diverse array of languages \citep{Scao22, Hoffmann22, Brown20}. In some scenarios, the translations generated by these models showcase such remarkable quality that they present a formidable challenge to established state-of-the-art translation models \citep{Hendy23, Zhang23}.

The task in which we performed our experiments and compared different models is that of \ac{imt}. This field has been under investigation since \citet{Foster97}, with the first appearance of the prefix approach, and has continued evolving ever since. Numerous research branches have emerged, exploring different techniques to help reduce the human effort during the translation sessions. \citet{Domingo17} introduced the segment-based protocol used in this paper, in which the user can validate multiple segments within a single translation. Other techniques directly target the minimization of human effort, such as optimizing the utilization of user mouse actions \citep{Navarro21b, Sanchis08a} or implementing a confidence measurement system for an initial evaluation of the translation quality \citep{Navarro21a, Gonzalez10a}. Additional strategies leverage the inherent capability of \ac{imt} systems to ensure perfect translations, utilizing them to enhance the translation model through active and online learning techniques \citep{Peris19, Peris18a, Gonzalez14}. Comprehensive frameworks like PRIMT \citep{Cheng16}, Casmacat \citep{Alabau13b}, TransType \citep{Cubel03} amalgamate a broad spectrum of these innovations within a unified workspace. Commercial environments incorporate interactive machine translation functionality, exemplified by platforms like \textit{Lilt} and \textit{Unbabel}.

In the upcoming section, we will explore the framework of segment-based \ac{imt}, which will provide a deeper understanding of how we have tailored the pre-trained models for this specific task.

\section{Segment Based IMT Framework}
To fully understand the behavior of the \ac{imt} system, it is essential to look at the \ac{nmt} framework introduced by \citet{Castano97}. This framework has consistently demonstrated its effectiveness in recent years, contributing to the advancements in the field of \ac{mt}, and it is widely used today \citep{Hendy23,Stahlberg20}. \Ac{nmt} aims to find the best translation $\hat{y}_1^{\hat{I}}=\hat{y}_1,...,\hat{y}_{\hat{I}}$ of length $\hat{I}$, given a source sentence $x_1^J = x_1,...,x_J$ of length $J$. The fundamental equation to this approach would be:
\begin{flalign} \label{eq:1}
	\hat{y}_1^{\hat{I}} &= \argmax{I, y_1^I} \Pr(y_1^I\ |\ x_1^J)         && \\\nonumber 
	&= \argmax{I, y_1^I} \prod_{i=1}^{I} \Pr(y_i\ |\ y_1^{i-1}, x_1^J)    &&
\end{flalign}
where $ \Pr(y_i\ |\ y_1^{i-1}, x_1^J)$ is the probability distribution of the next word given the source sentence $x_1^J$ and the previous words from the translation $y_1^{i-1}$.

In the Segment-Based \ac{imt} protocol, the human cooperates with the \ac{nmt} system to translate the source sentence $x_1^J$. This cooperation starts with the \ac{nmt} system proposing an initial translation and then the user providing feedback to the system. The user can validate segments to be kept in future iterations and correct an incorrect word. With the translation generated by the \ac{nmt} system, the user validates all correct segments and introduces a word correction. The actions performed by the user become the feedback information $f_1^N=f_1,...,f_N$ that the system will receive, where $f_1^N$ is a sequence of the $N$ non-overlapping validated segments, including the word correction performed inserted as a one-word segment. In response to this user feedback, the system generates a sequence of non-validated segments $\tilde{g}_1^N = \tilde{g}_1,...,\tilde{g}_N$ where each $\tilde{g}_n$ is a subsequence of words in the target language. This sequence complements the user's feedback to conform to the new hypothesis:
\begin{flalign}
	{y'}_1^I = f_1, \tilde{g}_1, ..., f_N, \tilde{g}_N
\end{flalign}
The iterative process of generating new hypotheses continues until the user validates the whole translation and does not perform any word correction. The \ac{imt} system aims to obtain a sequence of non-validated segments at each iteration that, combined with the user feedback, improve the translation quality.

The segments from the user feedback $f_1^N$ and those that are non-validated $\tilde{g}_1^N$ do not overlap, so the words produced by the system exclusively belong to either a validated or non-validated segment. The word generation process can be differentiated based on whether the words belong to a validated or a non-validated segment.

\citet{Peris17} formalized the word probability expression for the words belonging to the $n$-th validated segment $f_n$, that starts at the position $i_n+1$ of the translation, as:
\begin{flalign}
	& p(y_{i_n+i'} \mid y_1^{i_n+i'-1}, x_1^J, f_1^N) =   \\\nonumber
	& = p(y_{i_n+i'} \mid y_1^{i_n+i'-1}, x_1^J, f_1^N) * \delta(y_{i_n+i'}, f_{ni'})  \\\nonumber
	&& \hidewidth 1 \le i'\le \lvert f_n \rvert 	
\end{flalign}
where $\lvert f_n \rvert$ is the length of the validated segment $f_n$, $f_{ni'}$ refers to the $i'$-th word of such segment, and $\delta( \cdot , \cdot )$ is the Kronecker delta.

It is more complicated for the system to compute the words belonging to the non-validated segments as we do not know their length. To estimate the most probable non-validated segment and its length, we compute $M+1$ alternatives, each with a different length value. $M$ is the maximum length of the non-validated segments and is calculated on a development set. We set the $n$-th non-validated segment $\tilde{g}_n$ as the sequence of words that obtains the maximum probability, which is formulated as follows: 
\begin{flalign}
	\hat{g}_{n},_1^{\widehat{W}_n} = \argmax{0 \leq W_n \leq M \\ g_{n},_1^{W_n}} \prod_{i'=i_n+1}^{i_n+1+W_n} \Pr( y_{i'} \mid y_{1}^{i'-1}, x_1^J)
\end{flalign}
as before, $i_n+1$ is the first position of the segment $\tilde{g}_n$.

Note that this equation allows the system to set the length of the non-validated segment to zero for those cases where the maximum probability is obtained by merging two validated segments.

We have implemented a framework compatible with different models from the Hugging Face library \citep{Wolf20}, from where we obtained the mT5 \citep{Xue20} and mBART \citep{Liu20} model checkpoints. In the \ac{imt} system we developed, the \ac{nmt} model interacts with a simulated user to generate the translations. The simulated user compares the hypothesis generated by the system with the translation reference word-by-word to, in the first step, validate all the correct segments and, secondly, correct the first error found. This process is repeated until the source sentence is correctly translated. \Cref{sec:simulation} of the paper describes the simulation process in more detail.

\section{Experimental Framework}
This section offers a look at the experimental procedures undertaken in this paper. First, we explain the evaluation metrics employed to compare the systems and assess their quality in the \ac{imt} task. Then, we describe the corpora employed for the experiments along the different pairs of languages and sets used. Further, we sketch the specific fine-tuning procedures followed to develop our \ac{mt} systems. Finally, we describe in detail the interactive process of the system and how we have simulated the user to compute the results.

\subsection{Evaluation metrics}
We have utilized various well-known metrics to assess the quality of translations generated by the \ac{mt} models and the human effort involved when used within a segmented-based \ac{imt} system.

To assess the quality of the first hypothesis that each \ac{mt} system produces, we used the following metrics:

\begin{description}
	\item[\Ac{bleu}] \cite{Papineni02}: computes the geometric average of the modified $n$-gram precision, multiplied by a brevity factor that penalizes short sentences. In order to ensure consistent BLEU scores, we used \emph{sacreBLEU} \cite{Post18} for computing this metric.
	\item[\Ac{ter}] \cite{Snover06}: computes the number of word edit operations (insertion, substitution, deletion and swapping), normalized by the number of words in the final translation. It can be seen as a simplification of the user effort of correcting a translation hypothesis on a classical post-editing scenario.
\end{description}
Moreover, to assess the human effort involved in the interactive translation process, we used two metrics that differentiate between the effort performed with the keyboard and the mouse.

\begin{description}
	\item[\Ac{wsr}] \citep{Tomas06}: measures the number of words typed by the user, normalized by the number of words in the final translation.
	
	\item[\Ac{ksr}] \citep{Tomas06}: measures the number of characters typed by the user, normalized by the number of character in the final translation.

	\item[\Ac{mar}] \citep{Barrachina09}: measures the number of mouse actions made by the user, normalized by the number of characters in the final translation. 
\end{description}
The lower the value of these metrics, the better the system operates with the \ac{mt} model, as fewer interactions are needed to obtain perfect translations. It is noteworthy that \ac{wsr} and \ac{ksr} measure user interaction through the keyboard. The disparity between these metrics allows us to discern whether the system encounters greater difficulty with more complex and lengthier words. Additionally, for comparing the results, we give greater importance to keyboard effort, as some systems enable the automation of mouse interactions or a shift in the device employed for system navigation.

\subsection{Corpora}
In our experiments, we employed the Europarl corpus \citep{Koehn05}, a compilation of proceedings from the European Parliament. We used the training set of this corpora to fine-tune mBART and mT5 to the corpus domain. To validate and test the De{\textendash}En and Fr{\textendash}En models, we used WMT\footnote{\url{http://www.statmt.org/wmt12/translation-task.html}.}\footnote{\url{http://www.statmt.org/wmt15/translation-task.html}.}'s's \emph{news-test2013} and \emph{news-test2015} datasets, respectively. For the Es{\textendash}En models, we used \emph{news-test2012} and \emph{news-test2013} for validation and test purposes. It is worth noting that these datasets are commonly used in machine translation research and provide a benchmark for evaluating the performance of the models.

\cref{ta:corp} shows the main features of the corpora. 

\begin{table}
	\centering
	\resizebox{\columnwidth}{!}{
	\centering
	\begin{tabular}{c c c c c c}
		\toprule
		& & \multicolumn{3}{c}{\textbf{Europarl}} \\
		\cmidrule(lr){3-5}
		& & \textbf{{De{\textendash}En}} & \textbf{{Es{\textendash}En}} & \textbf{{Fr{\textendash}En}} \\
		\midrule
		\multirow{3}{*}{Train} 
		& $|S|$ & 1.9M 			& 2.0M 			& 2.0M 			\\
		& $|T|$ & 49.8M/52.3M 	& 51.6M/49.2M 	& 60.5M/54.5M 	\\
		& $|V|$ & 394.6K/129.1K	& 422.6K/309.0K	& 160.0K/131.2K \\
		\midrule
		\multirow{3}{*}{Val.} 
		& $|S|$ & 3000 			& 3003 			& 3000  		\\
		& $|T|$ & 63.5K/64.8K 	& 69.5K/63.8K 	& 73.7K/64.8K  	\\
		& $|V|$ & 12.7K/9.7K 	& 16.5K/14.3K 	& 11.5K/9.7K  	\\
		\midrule
		\multirow{3}{*}{Test} 
		& $|S|$ & 2169 			& 3000 			& 1500 			\\
		& $|T|$ & 44.1K/46.8K 	& 62.0K/56.1K 	& 29.9K/27.2K 	\\
		& $|V|$ & 10.0K/8.1K 	& 15.2K/13.3K 	& 6.3K/5.6K 	\\
		\bottomrule
	\end{tabular}}
	\caption{Corpora statistics. K denotes thousands and M millions. \emph{$|$S$|$} stands for number of sentences, \emph{$|$T$|$} for number of tokens and \emph{$|$V$|$} for size of the vocabulary. \textbf{{Fr}} denotes French; \textbf{{En}}, English; \textbf{{De}}, German; and \textbf{{Es}}, Spanish.}
	\label{ta:corp}
\end{table}

\begin{figure*}
	\footnotesize
	\centering
    \begin{minipage}{\textwidth}
    \centering
	\begin{tabular}{lll}
		\textbf{SOURCE}: & \multicolumn{2}{l}{El Estado de Indiana fue el primero en exigirlo.} \\
		\textbf{TARGET}: & \multicolumn{2}{l}{Indiana was the first State to impose such a requirement.} \\
		& & \\
		\multicolumn{1}{l|}{ \textbf{ITER-0}} & \multicolumn{1}{l|}{Translation hypothesis}   
		& Indiana is the sooner State to impose that condition. \\ \hline

		\multirow{2}{*}{\textbf{ITER-1}} & \multicolumn{1}{|l|}{Feedback} 
		& \val{Indiana} \err{was} \phantom{the sooner} \val{State to impose} \\
		 & \multicolumn{1}{|l|}{Translation hypothesis} 
		& \val{Indiana was} the sooner \val{State to impose} such a condition. \\ \hline

		\multirow{2}{*}{\textbf{ITER-2}} & \multicolumn{1}{|l|}{Feedback}    
		& \phantom{Indiana was} \val{the} \err{first} \phantom{State to impose} \val{such a} \\
		& \multicolumn{1}{|l|}{Translation hypothesis} 
		& \val{Indiana was the} \val{first} \val{State to impose} \val{such a} prerequisite. \\ \hline

		\multirow{2}{*}{\textbf{ITER-3}} & \multicolumn{1}{|l|}{Feedback}    
		& \phantom{Indiana was the first State to impose such a} \err{requirement} \\
		& \multicolumn{1}{|l|}{Translation hypothesis} 
		& \val{Indiana was the first State to impose such a requirement.} \\ \hline

		\multicolumn{1}{l|}{\textbf{END}}  & \multicolumn{1}{l|}{Final translation} 
		& \val{Indiana was the first State to impose such a requirement.} \\
        & & \\
        \multicolumn{2}{r}{\textbf{Post-editing effort:}} & 16 keystrokes and 8 mouse actions. \\
	\end{tabular}
	\end{minipage}
	\caption{Segment-based \ac{imt} session to translate a sentence from Spanish to English. The process starts with the system offering an initial hypothesis. Then, at iteration 1, the user validates the word segments \val{Indiana} and \val{State to impose} and makes a word correction (\err{was}). The system reacts to this feedback by generating a new translation hypothesis. Once more, the user reviews the hypothesis, validating this time the word segments \val{the} and \val{such a} and making the word correction \err{first}. In the third iteration, there are no more segments to validate, and the user only performs the word correction (\err{requirement}). Finally, since the next hypothesis is the desired translation, the process ends with the user accepting the translation. Overall, this process has a post-editing effort of 3 wordstrokes and 10 mouse actions.}
	\label{fig:ex_word}
\end{figure*}

\subsection{Systems}
Both checkpoint models, mBART and mT5, have been obtained through the Hugging Face library. Before using them, we had performed a fine-tuning for each pair of languages of the corpora, utilizing the training set. 

For mBART, we utilized the \textit{facebook/mbart-large-50-many-to-many-mmt}\footnote{\url{https://huggingface.co/facebook/mbart-large-50-many-to-many-mmt}} checkpoint \citep{Tang20} from the Hugging Face library \citep{Wolf20}, which employs a Seq2Seq Transformer architecture \citep{Vaswani17}. The model consists of $12$ encoder layers and $12$ decoder layers, with a model dimension of $1024$ and $16$ heads. For customizing the mBART model to a specific pair of languages of our domain, we fine-tuned it on a single bi-text dataset from the training set of the corpora. During training, we inputted the source language into the encoder and used the decoder to decode the target language, resulting in a new model for each language pair. We conducted 100K training updates with a learning rate of $2e-5$ and a weight decay of $0.01$ for training each model.

To obtain the mT5 models, we use the \textit{google/mt5-base}\footnote{\url{https://huggingface.co/google/mt5-base}} checkpoint \citep{Xue20} from the Hugging Face library \citep{Wolf20}, which also employes a Seq2Seq Transformer architecture \citep{Vaswani17}. The model consists of $12$ encoder layer and $12$ decoder layers, with a model dimension of $768$ and $12$ heads. For the fine-tuning process, we followed the same steps used for mBART. We have fine-tuned the model for each specific pair of languages for 100K training updates, with a learning rate of $2e-5$ and a weight decay of $0.01$.

For the \ac{sota} comparison of our systems, we have used the results obtained in the paper by \citet{Navarro23a}, where they trained an \ac{nmt} model and used it for \ac{imt} translation at the segment level. Their model is trained using the OpenNMT-py toolkit \citep{Klein17} with the Transformer architecture \citep{Vaswani17}. This model consists of $6$ encoder and decoder layers, with a model dimension of $512$ and $8$ heads.

\begin{table*}[!t]
	\footnotesize
	\resizebox{0.9\textwidth}{!}{\begin{minipage}{\textwidth}
    \centering
	\begin{tabular}{c c c c c c c c c}
		\toprule
		 &  & \multicolumn{2}{c}{\textbf{Translation Quality}} & \multicolumn{3}{c}{\textbf{User Effort}}\\
		\cmidrule(lr){3-4}\cmidrule(lr){5-7}
		\textbf{Model} & \textbf{Language Pair} & \textbf{TER [$\downarrow$]} & \textbf{BLEU [$\uparrow$]} & \textbf{WSR [$\downarrow$]} & \textbf{KSR [$\downarrow$]} & \textbf{MAR [$\downarrow$]} \\
		\midrule
        \multirow{6}{*}{mT5} 
        & De{\textendash}En & $69.3$ & $15.1$ & $52.17$ & $64.84$ & $19.78$\\
        & En{\textendash}De & $74.2$ & $13.2$ & $63.64$ & $66.00$ & $17.00$\\
		& Es{\textendash}En & $65.3$ & $18.1$ & $44.06$ & $51.49$ & $14.40$\\
        & En{\textendash}Es & $64.3$ & $18.4$ & $46.45$ & $55.57$ & $13.92$\\
		& Fr{\textendash}En & $66.3$ & $18.6$ & $44.74$ & $52.62$ & $15.02$\\
        & En{\textendash}Fr & $81.8$ & $17.8$ & $48.34$ & $55.73$ & $15.16$\\
		\midrule
		\multirow{6}{*}{mBART}
		& De{\textendash}En & $52.4$ & $29.7$ & $52.17$ & $68.13$ & $19.78$\\ 
        & En{\textendash}De & $57.0$ & $27.1$ & $50.00$ & $56.00$ & $14.00$\\ 
		& Es{\textendash}En & $52.1$ & $30.5$ & $33.08$ & $38.95$ & $12.19$\\ 
        & En{\textendash}Es & $48.2$ & $33.3$ & $34.41$ & $41.09$ & $11.68$\\
		& Fr{\textendash}En & $48.4$ & $33.6$ & $32.35$ & $37.90$ & $12.35$\\
        & En{\textendash}Fr & $56.0$ & $39.1$ & $29.92$ & $34.38$ & $11.07$\\
		\bottomrule
	\end{tabular}
	\end{minipage}}
	\caption{Results of the mT5, and mBART models in a segment-based \ac{imt} system. All values are reported as percentages.}
	\label{ta:res}
\end{table*}

\subsection{Simulation}\label{sec:simulation}
In order to mitigate the substantial time and financial expenses linked with recorrect human evaluations during the development phase, we chose to employ simulated users to perform the experiments and assess the models. Moreover, we opted for a more controlled experimental setting using simulations to minimize the number of possible external errors, removing the human factor of the equation. These simulated users were tasked with generating perfect translations from a given reference by providing feedback to the \ac{imt} system.

To perform these evaluations, we utilized the segment-based protocol described by \citet{Domingo17}, in which the user performs two actions to provide feedback to the system. First, he validates all the correct segments of the sentence and then performs a word correction that is treated as a one-word validated segment. With this information, the system generates a new hypothesis in which all the validated segments are present.

For the sake of simplicity and to maintain generality, in this simulation, we assumed that the user consistently corrects the leftmost incorrect word. Additionally, we stipulated that validated word segments must follow the same order as in the reference. This assumption aligns with the approach adopted by the original creators of the segment-based protocol \citep{Domingo17,Peris17}.

When the simulation starts, the system generates an initial hypothesis for the translation, which the simulated user then reviews. First, the user must validate all the correct word segments obtained by computing the longest common subsequence \citep{Apostolico87} between hypothesis and reference. This action has an associated cost of one for each one-word segment and two for each multi-word segment, simulating the click-and-drag actions from the mouse for selecting the validated segments. Then, the user looks for pairs of consecutive validated segments that appear consecutively in the reference. For each segment merge, the number of mouse actions increases by one if there is a single word between them or two otherwise. Again, this simulated the click-and-drag actions to delete those errors. Finally, the user corrects the leftmost incorrect word. This action costs one mouse action for mouse movement to the error position and one keystroke per character typed. Then, the system reacts to this feedback by generating a new hypothesis. This process is repeated until the hypothesis and the reference are the same.

\Cref{fig:ex_word} presents an example of the simulation performed to translate a source sentence. The translation session starts with the system generating an initial hypothesis that needs to be reviewed and corrected. Then, at iteration 1, the user validates two segments by performing three mouse actions and corrects the word \emph{was}. Note that the first validated segment is just one word, so the user only needs one mouse action to validate it. At iteration 2, the simulated user repeats the process, validates two new segments, and performs the word correction \emph{first}. The total cost of the iteration is three mouse actions and one wordstroke. There are no correct segments to validate at the third iteration, so the user just performs the word correction. Finally, the translation hypothesis that the system generates with the new feedback is correct, and the simulated user validates it with a total effort of three wordstrokes and ten mouse actions.

\section{Results}
In order to perform the comparison of the pre-trained models, we evaluated in a segment-based IMT system the models mT5 and mBART. We expect to see a significant reduction in the human effort using this method, as mBART has already demonstrated in a previous paper that can obtain state-of-the-art results on a prefix-based IMT system. In this case, we evaluate the human effort with three different metrics, the WSR, the KSR, and the MAR, each evaluating a different kind of effort. Both WSR and KSR measure the effort performed by typing the words, but as KSR is calculated at the character level, it gives more importance to longer words. The MAR metric evaluates the effort performed by clicking and moving the mouse. In order to achieve a more precise evaluation of the models, we consider that the effort required to type a word exceeds that of moving the mouse. Also, in a professional environment with real translators, there are ways to perform the mouse actions with the keyboard to evade moving the hands from there, diminishing their implication.

\Cref{ta:res} shows the experimental results, where the mT5 models are compared with mBART. mBART achieves superior translation quality for all language pairs compared to mT5, consistently achieving higher BLEU and TER scores in all cases. These outcomes are initially logical, considering that mBART has been pre-trained for translation tasks, allowing fine-tuning to specialize its domain for the specific task. On the contrary, mT5 is not pre-trained for any specific task; in fact, using the model requires fine-tuning for the particular task.

Given the disparity in the initial translation quality between the two models, it can be inferred that mBART is more effective in reducing the human effort required for the translation task, a deduction substantiated by empirical evidence. However, it is noteworthy that mT5, achieving a BLEU score of around 17 points, only needs the user to type 50\% of the translation words. This outcome, while commendable compared to translating from scratch, is less impactful in the field of IMT.

Conversely, mBART, with a BLEU score of around 30 points, attains a \ac{wsr} of 30 points, approximating the state-of-the-art in the field, except for the De{\textendash}En language pair. To facilitate a comprehensive benchmark against the state-of-the-art in segment-level \ac{imt}, we intend to compare mBART with results obtained from a previous study within the same IMT field \citep{Navarro23b}.

\begin{table*}[!t]
	\footnotesize
	\resizebox{0.9\textwidth}{!}{\begin{minipage}{\textwidth}
    \centering
	\begin{tabular}{c c c c c c c c}
		\toprule
		 &  & \multicolumn{2}{c}{\textbf{Translation Quality}} & \multicolumn{2}{c}{\textbf{User Effort}}\\
		\cmidrule(lr){3-4}\cmidrule(lr){5-6}
		\textbf{Model} & \textbf{Language Pair} & \textbf{TER [$\downarrow$]} & \textbf{BLEU [$\uparrow$]} & \textbf{KSR [$\downarrow$]} & \textbf{MAR [$\downarrow$]} \\
		\midrule
        \multirow{4}{*}{OpenNMT-py \citep{Navarro23b}} 
        & De{\textendash}En & $56.4$ & $24.7$ & $28.7$ & $27.8$\\
        & En{\textendash}De & $60.2$ & $21.9$ & $29.8$ & $23.3$\\
		& Es{\textendash}En & $55.4$ & $26.8$ & $27.0$ & $27.4$\\
        & En{\textendash}Es & $53.0$ & $28.3$ & $27.7$ & $26.1$\\
		\midrule
		\multirow{4}{*}{mBART}
		& De{\textendash}En & $52.4$ & $29.7$ & $68.13$ & $19.78$\\ 
        & En{\textendash}De & $57.0$ & $27.1$ & $56.00$ & $14.00$\\ 
		& Es{\textendash}En & $52.1$ & $30.5$ & $38.95$ & $12.19$\\ 
        & En{\textendash}Es & $48.2$ & $33.3$ & $41.09$ & $11.68$\\
		\bottomrule
	\end{tabular}
	\end{minipage}}
	\caption{Results of the \emph{OpenNMT-py}, and mBART models in a segment-based \ac{imt} system. All values are reported as percentages.}
	\label{ta:res2}
\end{table*}

\Cref{ta:res2} contains the experiment results comparing mBART with an NMT model from a previous paper using segment-based IMT. In this case, we do not have the pairs of languages Fr-En and vice versa, and we solely utilize the KSR metric to measure the reduction in human effort. In this scenario, both models achieve a more comparable translation quality, but it is here where pre-trained models shine the most. We obtained better translations with a model that we did not have to train from scratch, reducing the time and computation cost, and it required less data for fine-tuning. Although it does not reach the reduction in effort achieved by the OpenNMT model, the values for language pairs involving Spanish are very similar.

If we examine the translation process of mT5 and mBART, we can see that we came to the same conclusion as to why they have not achieved results as good as those of the model trained from scratch. Despite both models producing valid and correct translations in the initial iteration, these are equal to the reference, and the simulated user expects to receive it, initiating the iterative process to reach it. At this point, pre-trained models encounter the most challenges; they cannot adapt and generalize as smoothly as the system trained from scratch. The non-validated segments generated by the system to fill the gaps between those validated are of lower quality, leading the user to write more words. Hence, the MAR metric value is relatively low in both cases because new segments are not validated as frequently as in the OpenNMT model, so the mouse is used less frequently.

In summary, the best results in terms of KSR for reducing the human effort during interactive machine translation sessions in a segment-based environment have been achieved with the model trained from scratch. The main reason is that the pre-trained models have not been able to generalize the non-validated segments well. Although pre-trained models have obtained state-of-the-art results in a large set of translation tasks, they seem to have problems generalizing words between two validated segments. 

\section{Conclusions and future work}
This study has compared the effectiveness of pretrained multilingual machine translation models. mBART has obtained better effort reduction than mT5 for all the pairs of languages, so in the case of having to choose one of them, mBART is the best solution. The model has been trained initially to perform machine translation, and we can obtain high-quality translations by fine-tuning it for the specific domain. mBART has obtained results similar to those of an NMT model trained from scratch. We can obtain equivalent performance in a segment-based IMT system, significantly reducing the computational cost associated with training. 

In future work, it would be interesting to investigate the behavior of other pretrained models in the IMT field and try to use LLMs by performing prompting engineering or working with 0-shot, few-shot techniques. Additionally, conducting a comparative analysis among these models would provide valuable insights.

\section*{Acknowledgements}
\phantom{This work received funding from \emph{Generalitat Valencia} under the program \emph{CIACIF/2021/292} and from \emph{ValgrAI} (\emph{Valencian Graduate School and Research Network for Artificial Intelligence}). Work partially supported by grant PID2021-124719OB-I00 funded by MCIN/AEI/10.13039/501100011033 and by \emph{European Regional Development Fund} (\emph{ERDF}).}

\end{document}